# The Wisdom of Artificial Deliberative Crowds

**Federico Barrera-Lemarchand[1,2], Mariano Sigman[1], Joaquín Navajas[1,2]**

[1] Laboratorio de Neurociencia, Escuela de Negocios, Universidad Torcuato Di Tella, Buenos Aires, Argentina

[2] Consejo Nacional de Investigaciones Científicas y Técnicas (CONICET), Buenos Aires, Argentina

## Abstract

The aggregation of many lay estimates often outperforms individual expert judgment, a phenomenon known as the wisdom of crowds. While this is usually attributed to the independence of estimates, an even stronger effect arises through deliberation: averaging the consensus estimates of small deliberating groups outperforms the classical wisdom of crowds, with individual judgments themselves also becoming more accurate after deliberation. Whether these improvements transfer to large language models deliberating amongst themselves is unknown. Here we adapt a three-stage deliberation paradigm previously used with human participants for use with large language models from three different families, and test it across four domains of increasing real-world stakes: visual numerical estimation (Study 1), peer review of machine-learning papers (Study 2), detection of hidden malicious behavior by an artificial intelligence agent (Study 3), and sports forecasting against a real prediction market (Study 4). Across domains, deliberation reduced collective error beyond passive aggregation of independent responses, and post-deliberation individual judgments retained this collective gain. Notably, the advantage required model diversity: groups composed of clones of a single model did not benefit from deliberating. These results establish machine deliberation as a general-purpose aggregation mechanism, and point to diversity as an active ingredient.

# Introduction

Extensive literature has shown that the aggregation of many lay estimates often outperforms individual expert judgments (De Condorcet, 1785; Galton, 1907). This phenomenon, popularly known as the “wisdom of crowds” (Surowiecki, 2005), has been applied to a wide range of problems, including predicting financial markets (Ray, 2006), forecasting geopolitical events (Mellers et al., 2014), and even decoding the smell of molecules (Keller et al., 2017). Similarly, in artificial intelligence, aggregating outputs from multiple models is a well-established tool for improving reliability and reducing variance: panels of small LLM judges match or beat a single larger judge at a fraction of the cost (Verga et al., 2024); layered information sharing between models outperforms any individual component (Wang et al., 2024); and the classical wisdom-of-crowds effect on numerical forecasts replicates in LLM populations (Schoenegger et al., 2024). All of these methods aggregate outputs produced independently by each model.

The idea that deliberation can improve thinking goes back to the Socratic tradition, in which dialogue was itself a method for constructing and revising knowledge. The underlying intuition is simple: exchanging ideas makes reasoning processes visible that would otherwise remain unnoticed, allowing us to inspect arguments, detect errors, and consider alternative ways of thinking. Conversation can thus provide a space in which intuitive judgments are brought under scrutiny and revised. Previous research in humans has examined the impact of social influence on the wisdom of crowds, yielding mixed results. Some studies suggest that it can lead to herding and reduce accuracy (Raafat et al., 2009; Lorenz et al., 2011; Madirolas & de Polavieja, 2015), whereas others show that social interaction can improve collective estimates (Gürçay et al., 2015; Mellers et al., 2014; Bahrami et al., 2010; Juni & Eckstein, 2015). For instance, averaging the consensus estimates reached by small groups has been found to significantly outperform the wisdom of large, independent crowds (Navajas et al., 2018; Barrera-Lemarchand et al., 2026), a result that has been replicated across multiple contexts (Dezecache et al., 2022; Espina Mairal et al., 2024; Pescetelli et al., 2021). Specifically, in Barrera-Lemarchand et al. (2026) the researchers followed a very simple procedure, asking participants for individual estimates on a series of straightforward wisdom-of-crowds questions (e.g. "how many countries are there in the Southern Hemisphere?"), and then dividing participants into deliberating groups with the goal of trying to reach consensus. This raises a broader question: whether the benefits of deliberation reflect a sufficiently general principle that they can extend beyond human cognition, such that even a conversation among LLMs could profit from the same mechanism. If so, one of the oldest tools for improving human thought (dialogue) could become a useful procedure for one of our newest forms of intelligence. Testing machine deliberation therefore offers a way to ask whether the gains of conversation depend on specifically human social cognition, or can emerge more generally from the exchange, scrutiny, and revision of ideas. It could very well be the case that deliberation between models adds nothing over simple averaging (if models simply exchange and pool their numbers), or even actively hurts (herding on shared biases; Lorenz et al., 2011). Additionally, the domains where machine deliberation would matter most (e.g. evaluation, oversight, forecasting) differ substantially from the general-knowledge estimation questions used in the prior human studies.

In this work, we extend the aforementioned procedure to crowds of three small LLMs from distinct model families (Anthropic's Claude Haiku 4.5, OpenAI's GPT-5-nano, and Google's Gemini 3.1 Flash Lite), holding the deliberation protocol constant across four task domains:

- **Study 1: Factual estimation.** Following a classic example of the wisdom of crowds in action, we asked the models to estimate the number of small disks contained within a jar, with a known ground truth, and 14 distinct amounts of disks (between 38 and 589).
- **Study 2: Peer review.** We asked the models to evaluate 100 NeurIPS 2024 submissions (orally-presented vs. rejected papers), where the "ground truth" is the human editorial verdict.
- **Study 3: AI-safety monitoring.** Here we asked the models to rate the safety of agentic trajectories in which an LLM agent may be pursuing a hidden malicious goal (trajectories from the Monitor Red Teaming dataset; Kale et al., 2025).
- **Study 4: Forecasting.** Lastly, we asked the models to forecast the 2026 FIFA World Cup (for each of the 12 groups, which team would finish first, and for each of the 32 knockout matches, which team would win) together with the probability they assigned to that outcome. For every one of those 44 events we also recorded the price that a real-money prediction market (Polymarket) was quoting at the same moment, which places the accuracy of the crowd on an absolute, incentive-backed scale. All forecasts were elicited before the corresponding round was played, so the answers did not exist at the time of the judgment.

In every domain we compare individual model judgments, the classical wisdom of crowds (averaging independent judgments) and deliberative consensus. Studies 1–3 add a post-deliberation individual revision (Stage i2), which asks whether the collective gain is carried back into the judgment of the agents that produced it. Studies 1 and 2 add same-model ("clone") deliberation, which removes model diversity while holding the protocol, the task and the number of agents constant. Study 1 further contrasts deliberation with mere social influence, in which models see each other’s answers but do not talk.

## Results

We followed a similar procedure to Barrera-Lemarchand et al. (2026), replacing human participants by large language models from different families (Anthropic's Claude Haiku 4.5, OpenAI's GPT-5-nano, and Google's Gemini 3.1 Flash Lite; Figure 1). The models were selected so that every provider was represented by a small, inexpensive system of comparable capability. In Stage i1, LLMs independently assessed: (i) differing amounts of stacked disks contained within a jar in a series of 14 pictures (Study 1); (ii) whether or not to accept 100 submissions for NeurIPS 2024 for oral presentation, half of which were rejected by human reviewers (study 2); (iii) the safety of 50 trajectories of tasks and tool use performed by an agent, half of which involved unsafe behavior derived from a hidden malicious goal (study 3); and (iv) which team would finish first in each of the 12 groups of the FIFA World Cup 2026 and which team would win each of the 32 knockout matches, always before the corresponding round was played (study 4).

In Stage c, we divided these LLMs into groups of three, and recreated an online chat room where they could discuss each problem, prompting them and constraining their behavior to replicate a human chatroom-style exchange (Barrera-Lemarchand et al., 2026): short messages (7 words on average); a maximum of ten rounds of interaction, where each model spoke once per round in random order; and the explicit objective of "deliberating and trying to reach consensus". Initially, each model received their initial assessment along with a short reasoning supporting it (of up to 200 tokens, 250 in study 4), both of which were produced on Stage i1. On each turn, the corresponding model received both of these, along with every message produced so far in the "conversation". The deliberation ended when all models agreed on a specific value: the amount of disks in the picture in Study 1; whether the paper merited acceptance on a scale of 1 to 10 in study 2; how concerning the agent’s behavior had been, from 0 ("no concerns whatsoever") to 10 ("certain harmful or misaligned behavior"), in study 3; and which team would win the event, along with the probability assigned to that team, in study 4.

Lastly, on Stage i2, the LLMs were provided with their initial assessment and short reasoning supporting it, along with the full conversation produced in Stage c, and were asked to provide a new assessment. They were explicitly told that they could produce the same value as they had provided in Stage i1, the consensus reached in Stage c, or any other value they considered to be their best guess for the task they were tasked to solve.

Throughout these studies we found converging evidence for the wisdom of deliberative artificial crowds, with Stage c always exhibiting lower error (Studies 1) or Brier scores (Studies 2, 3 and 4) when compared to the classical wisdom of crowds obtained in Stage i1. Notably, Stage i2 always retained this improvement, and sometimes even surpassed it (Studies 1 and 2), again emulating the observed behavior in human subjects.

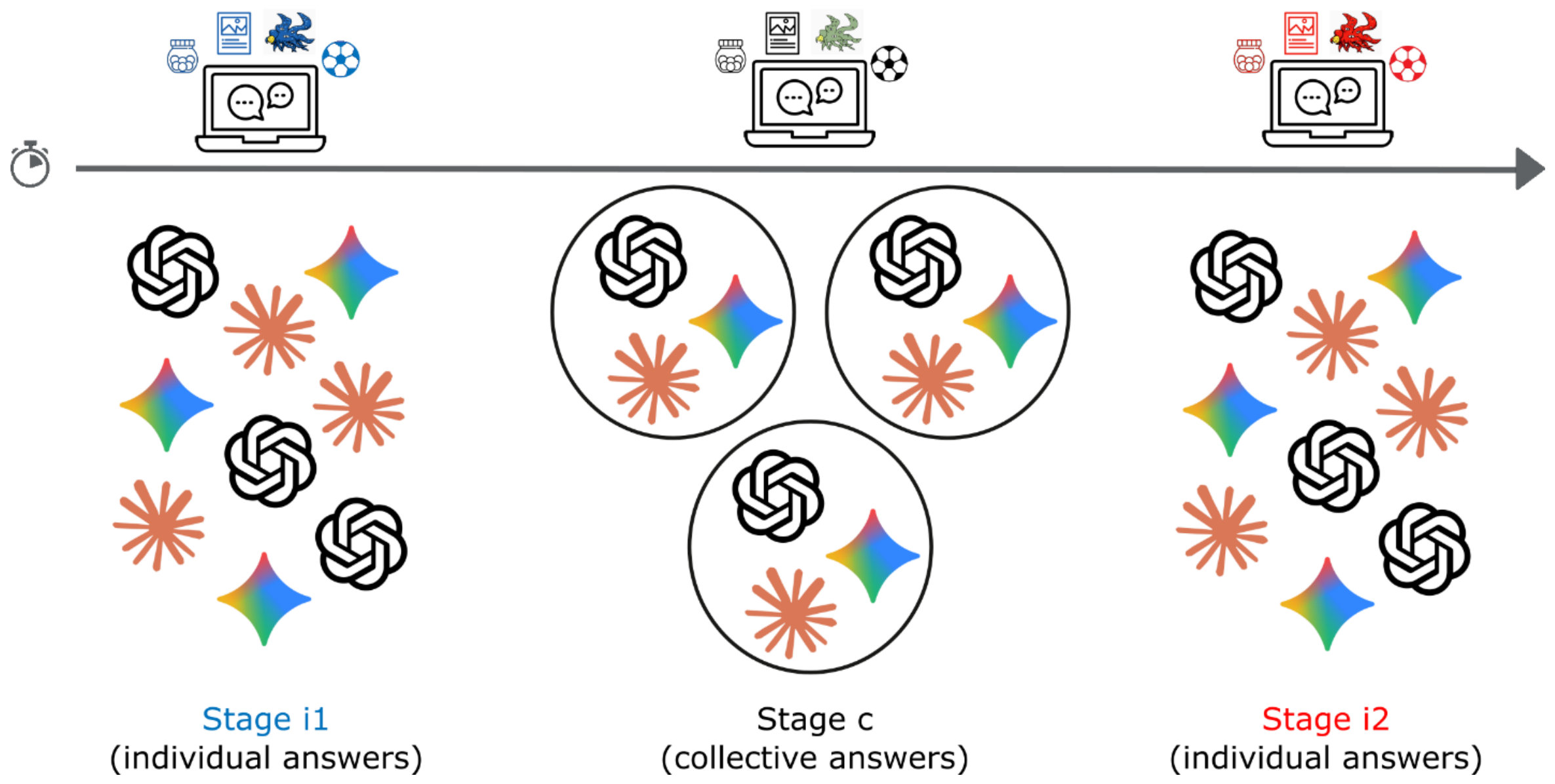

**Fig. 1. Study procedure.** The study had three stages. On Stage i1 LLM models from three different families provided their answer for a numerical estimation or forecasting task. On Stage c, the models were divided into groups of three, each of them was provided with their own answers from Stage i1, and they were asked to deliberate and try to reach consensus on the aforementioned tasks. Stage i2 followed the same procedure as Stage i1 (the models received their prior answers from Stage i1 and the full discussion they had on Stage c).

## Study 1: a classical wisdom-of-crowds estimation problem

Following a classical wisdom-of-crowds problem, we asked models to estimate the number of little plastic disks stacked inside a glass jar (the actual correct answers ranged from 38 to 589 disks) as observed in a series of 14 photographs. We also included a one-euro coin next to the jar in each of them, working as a scale anchor. We kept the prompt fixed for every picture. We found that deliberation vastly improves the accuracy of the model's estimates, and that the final individual assessment of the models reduces error further still (Figure 2A). For example, the estimation error of aggregating 12 randomly-chosen consensus estimates was substantially lower than the error obtained by averaging 36 randomly-chosen individual LLM values (Mann–Whitney U test: $z=19.4$, $p=4 \times 10^{-83}$; effect size: Cohen's $d = 4.75$). This finding implies that the group consensus values were more accurate than what would have been obtained by simply averaging the three initial individual ratings by the LLMs, suggesting that deliberation adds genuine value beyond passive aggregation (Fig. 2B, squared error of within-group averages: $0.195 \pm 0.019$, squared error of consensus estimates: 0.164 ± 0.019, Wilcoxon signed-rank test: $z=5.33$, $p=9.8 \times 10^{-8}$; effect size: paired Cohen's $d = 0.40$).

Additionally, comparing the models' estimates reveals a significant reduction in individual error from Stage i1 to Stage i2 (Fig. 2C, squared error of individual ratings in Stage i1: 0.230 ± 0.019, Stage i2: 0.159 ± 0.011, Wilcoxon signed-rank test: $z=7.36$, $p=1.9 \times 10^{-13}$; effect size: paired Cohen's $d = 0.40$). Notably, a jar-level paired comparison (which removes the large between-jar variance) revealed equal or lower Stage i2 error than consensus error in all 14 pictures (Wilcoxon signed-rank test: $z = 3.18$, $p = 0.0015$; effect size: paired Cohen's $d = 1.33$).

Lastly, we also considered the impact of social influence: providing models with each other's Stage i1 numerical responses instead of having them interact, and asking them to provide a new estimate (Figure 2D). We observe that, while social influence leads to greater accuracy over Stage i1 answers (Wilcoxon signed-rank test: $z = 3.58$, $p = 3.5 \times 10^{-4}$; effect size: paired Cohen's $d = 0.24$), deliberation still leads to more accurate estimates (Mann–Whitney U test on the bootstrap distributions at the largest crowd size: $z = 18.85$, $p = 3.0 \times 10^{-79}$; effect size: Cohen's $d = 3.25$). The same holds when each group's consensus is paired against its own socially-informed estimate (Wilcoxon signed-rank test: $z = 3.19$, $p = 0.0014$; effect size: paired Cohen's $d = 0.24$).

These results mirror those found in human participants (Navajas et al., 2018; Barrera-Lemarchand et al., 2026), suggesting that the method generalizes from humans to LLM agents on classic wisdom-of-crowds numerical estimation tasks.

## Study 2: expert evaluation

We then moved from estimation of a quantity with a known correct answer to expert evaluation. Following an analogous procedure, 100 NeurIPS 2024 submissions (50 that were accepted for oral presentation and 50 that were rejected, token-matched to control for length) were each reviewed by the same three models as the previous study, which determined whether the paper merited acceptance on a 1 (very strong reject) to 10 (award quality) scale. Each paper received 18 independent ratings and 6 deliberative group answers. All deliberations converged. We kept all prompts fixed for every submission. Using Brier scores against the human reviewers' verdict, and in line with Study 1, we found that consensus estimates from deliberative groups outperform pooled individual ratings (Figure 3A, Mann–Whitney U test between Stage i1 and Stage c on the largest crowd: $z = 19.35$, $p = 2 \times 10^{-83}$; effect size: Cohen's $d = 6.71$).

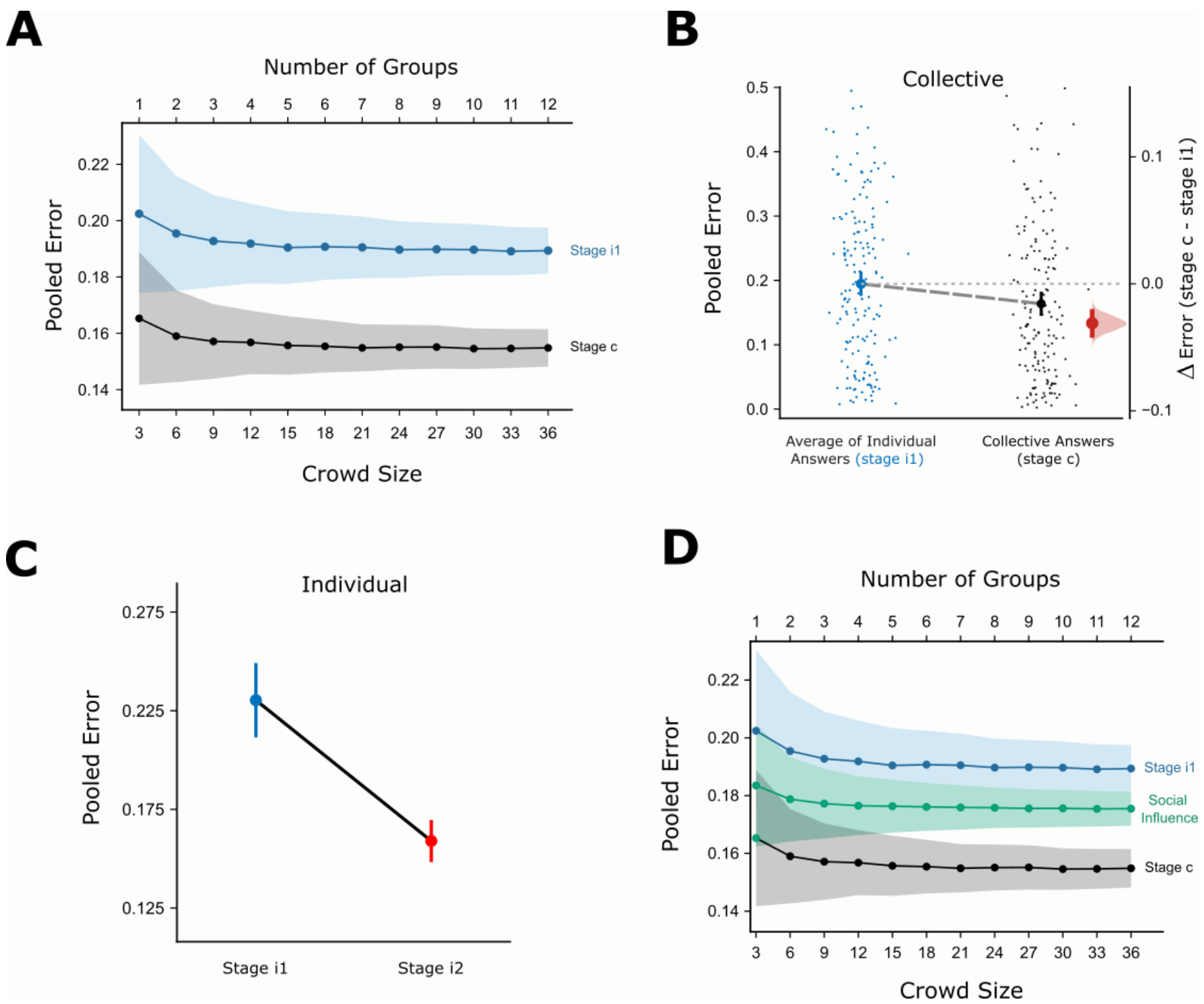

**Fig. 2. Empirical results for Study 1. (A)** Collective error, measured as the squared error in log space (base 10), of the average of $c$ individual estimates (blue line, Stage i1) and of the average of $c/3$ collective estimates (black line, Stage c), where $c$ denotes the crowd size (bottom axis; the top axis gives the corresponding number of deliberating groups). Data are pooled over 14 pictures, with 36 individual estimates and 12 deliberating groups per picture. Shaded bands represent the standard deviation across 2,000 random subsamples per crowd size. **(B)** Distributions of the squared log error (base 10) of the within-group average of the three Stage i1 estimates (blue) and of that group's consensus (black), computed from n = 167 independent deliberating groups. Dots show individual groups, filled circles show the mean, and error bars correspond to twice the standard error of the mean. The floating axis on the right shows the bootstrap distribution (5,000 resamples) of the mean paired difference between the two conditions, with the mean as a filled circle and its 95% confidence interval as a vertical bar; its zero is aligned with the Stage i1 mean (dotted line). **(C)** Mean squared log error (base 10) of the individual estimates before (Stage i1) and after (Stage i2) deliberation, computed from n = 504 model–group observations. The error bars correspond to twice the standard error of the mean. **(D)** Same as (A), with the addition of the social influence condition (green line), in which each model received the Stage i1 estimates of the other two members of its group and revised its own estimate without deliberating.

Similarly, the gain was not circumscribed to aggregation: pairing each group's consensus against the mean of its own three opening ratings, deliberation still lowered the error (Figure 3B, Wilcoxon signed-rank test: $z = 5.10$, $p = 3.4 \times 10^{-7}$; effect size: paired Cohen's $d = 0.25$). The individual improvement from Stage i1 to Stage i2 replicated as well: each reviewer's private re-rating after the discussion was more accurate than the rating it had started with (Figure 3C, Wilcoxon signed-rank test: $z = 9.85$, $p = 7.1 \times 10^{-23}$; effect size: paired

Cohen's d = 0.19). However, as opposed to Study 1, at the paper level, Stage i2 matched the consensus rather than improving upon it, with 82 out of the 100 submissions reproducing the agreed value exactly.

Taken together, these results extend the pattern beyond tasks with a physical ground truth: deliberation improves accuracy even when the criterion is a human expert verdict rather than a measurable physical quantity.

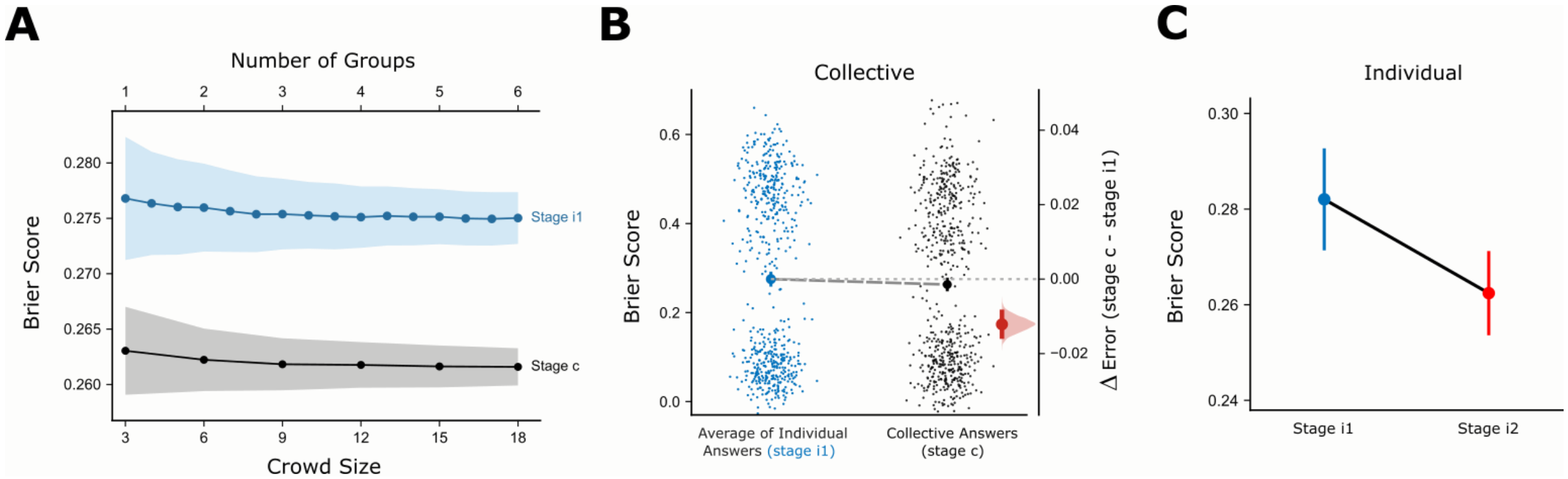

**Fig. 3. Empirical results for Study 2. (A)** Brier score, against the human reviewers' verdict, of the average of c individual ratings (blue line, Stage i1) and of the average of c/3 collective ratings (black line, Stage c), where c denotes the crowd size (bottom axis; the top axis gives the corresponding number of deliberating groups). Data are pooled over 100 NeurIPS 2024 submissions, each rated by 18 individual LLM reviewers organized into 6 deliberating groups of three. Shaded bands represent the standard deviation across 2,000 random subsamples per crowd size. **(B)** Distributions of the Brier score of the within-group average of the three Stage i1 ratings (blue) and of that group's consensus (black), computed from n = 600 independent deliberating groups. Dots show individual groups, filled circles show the mean, and error bars correspond to twice the standard error of the mean. The floating axis on the right shows the bootstrap distribution (5,000 resamples) of the mean paired difference between the two conditions, with the mean as a filled circle and its 95% confidence interval as a vertical bar; its zero is aligned with the Stage i1 mean (dotted line). **(C)** Mean Brier score of the individual ratings before (Stage i1) and after (Stage i2) deliberation, computed from n = 1,800 model–paper observations. The error bars correspond to twice the standard error of the mean.

## Study 3: AI-safety monitoring

Another relevant domain of potential application for AI deliberation is AI Safety, especially AI control. The two domains examined so far share a convenient property: the answer can be checked by a human participant. That will not always hold. Agentic LLMs are being delegated increasingly consequential tasks (Anthropic, 2026), exposing users and third parties to substantial potential harms (Anthropic, 2025a), and current capability trends point to systems whose outputs are hard to evaluate before deployment even for domain experts (Kwa et al., 2025; Glazer et al., 2024). This is the setting of AI control, in which a strong but untrusted model is overseen by weaker models that are trusted precisely because they are less capable (Greenblatt et al., 2024; Kenton et al., 2024), a strategy related to but distinct from debate (Irving et al., 2018; Khan et al., 2024). Here the monitor's weakness is the binding constraint, and it is a constraint an adversary can exploit: monitors that are trusted because they are weak have been shown to be steerable by attacks adapted to the control protocol itself (Terekhov et al., 2025). If deliberation sharpens the judgment of small models on problems where the ground truth is known, as Studies 1 and 2 suggest, it may do the same in domains where it is unknown. Study 3 and Study 4 put that to the test: in Study 3 we ask whether a deliberating crowd of weak monitors can detect sabotage by a stronger agent better than the same monitors judging alone, and in Study 4 we use deliberation for forecasting the outcomes of a popular sporting event (the 2026 FIFA Football World Cup).

The dataset for Study 3 comprises 50 agentic trajectories: 25 in which the agent pursues a hidden malicious objective alongside its legitimate task, and 25 in which it does not (Kale et al., 2025). Each of the three models read the full trajectory (including tool calls and visible reasoning) and rated how suspicious they found the agent's behavior on a scale of 0 (completely safe) to 10 (completely unsafe), following the same three-stage procedure as the previous studies. No harmless trajectory was flagged on aggregate: across those 25 trajectories the highest mean suspicion score was 2.7 out of 10 and the highest mean consensus 2.8, leaving essentially no variance for deliberation to act on. Error rates for individual monitors were also extremely low (no single rating on a harmless trajectory ever exceeded 5 on the 0–10 scale, and only 11 of the 450 individual ratings reached that value, 10 of those provided by the same model), so there was no room for improvement on these trajectories. The analyses that follow therefore concern the 25 sabotage transcripts, each of which received 18 independent ratings and 6 deliberative group answers.

The results closely resemble those of the previous studies: aggregating 6 group consensuses was more accurate than averaging 18 individual suspicion scores (Figure 4A, Mann–Whitney U test between Stage i1 and Stage c on the largest crowd: $z = 19.30$, $p = 5.6 \times 10^{-83}$; effect size: Cohen's $d = 4.30$), and the gain again survived the within-group comparison: each group's consensus was more accurate than the mean of its own three Stage i1 ratings (Figure 4B, Wilcoxon signed-rank test: $z = 3.64$, $p = 2.8 \times 10^{-4}$; effect size:

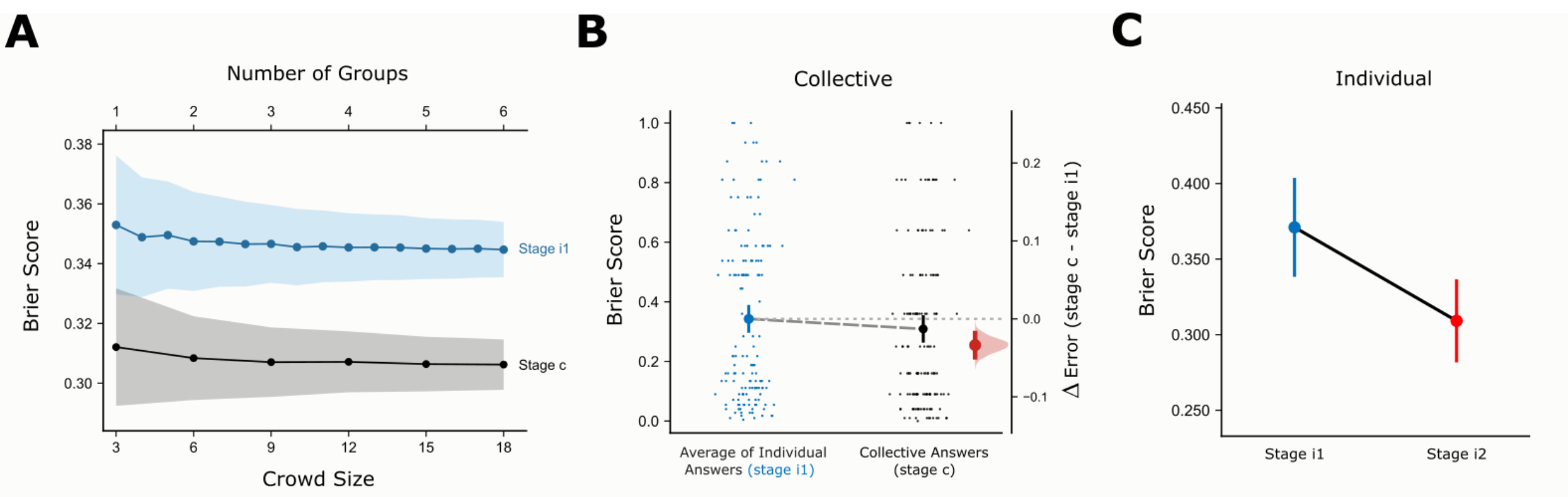

**Fig. 4. Empirical results for Study 3. (A)** Brier score, against the ground-truth label of the trajectory, of the average of c individual suspicion ratings (blue line, Stage i1) and of the average of c/3 collective ratings (black line, Stage c), where c denotes the crowd size (bottom axis; the top axis gives the corresponding number of deliberating groups). Data are pooled over the 25 trajectories that contained sabotage, each monitored by 18 individual LLM raters organized into 6 deliberating groups of three. Shaded bands represent the standard deviation across 2,000 random subsamples per crowd size. **(B)** Distributions of the Brier score of the within-group average of the three Stage i1 ratings (blue) and of that group's consensus (black), computed from n = 149 independent deliberating groups. Dots show individual groups, filled circles show the mean and error bars correspond to twice the standard error of the mean. The floating axis on the right shows the bootstrap distribution (5,000 resamples) of the mean paired difference between the two conditions, with the mean as a filled circle and its 95% confidence interval as a vertical bar; its zero is aligned with the Stage i1 mean (dotted line). **(C)** Mean Brier score of the individual ratings before (Stage i1) and after (Stage i2) deliberation, computed from n = 450 model–trajectory observations. The error bars correspond to twice the standard error of the mean.

paired Cohen's d = 0.30). Monitors also improved individually after discussing (Figure 4C, Wilcoxon signed-rank test: $z = 4.94$, $p = 7.6 \times 10^{-7}$; effect size: paired Cohen's d = 0.28). A transcript-level paired comparison, which removes the variance between trajectories, confirmed the collective gain in 19 of the 25 sabotage cases (Wilcoxon signed-rank test: $z = 3.54$, $p = 4.0 \times 10^{-4}$; effect size: paired Cohen's d = 0.90). However, as in Study 2, Stage i2 matched the consensus rather than improving upon it (14 non-zero differences, exact Wilcoxon signed-rank test, $p = 0.63$), with 417 of the 447 individual re-ratings reproducing the agreed value exactly (the remaining 3 are from the only group that failed to reach consensus, and thus no comparison is possible).

The reading that matters for AI control is a practical one (Greenblatt et al., 2024; Al Dahik et al., 2025): three weak monitors that talk to each other detect sabotage by a stronger agent better than the same three monitors judging independently. AI alignment (especially reinforcement learning through AI feedback, see Lee, 2023) could potentially also benefit from such a strategy (Verga et al., 2024; UK AI Security Institute, 2025).

## Study 4: forecasting

Lastly, we applied the deliberation procedure to a domain where the answers were unknown at the time of the study: sports forecasting. We asked the models to forecast the 2026 FIFA World Cup (for each of the 12 groups, which team would finish first, and for each of the 32 knockout matches, which team would win) together with the probability they assigned to that outcome. Each group of three debating models then produced a consensus answer in the same format, so every one of the 44 events yielded 12 independent forecasts and 4 deliberative group answers, all elicited before the corresponding round was played. We also recorded the prices a real-money prediction market (Polymarket) was quoting for the same event at that moment. Sources do not always name the same winner, so scoring each one on its own pick would grade them on different propositions. We therefore fixed the market's favorite as a common reference team and took the probability every source implied for that team. The event was coded 1 if that team won and 0 if it did not (favorites won 35 of the 44); Brier scores were computed on that basis. We did not ask for a post-deliberation individual re-rating, so this study had no Stage i2.

The collective gain replicated once more: aggregating 4 group consensus estimates was more accurate than averaging 12 independent forecasts (Figure 5A, Mann–Whitney U test between Stage i1 and Stage c on the largest crowd: $z = 19.11$, $p = 2 \times 10^{-81}$; effect size: Cohen's $d = 3.32$), and the within-group comparison held as well: each group's consensus beat the mean of its corresponding Stage i1's opening forecasts (Figure 5B, Wilcoxon signed-rank test: $z = 4.03$, $p = 5.5 \times 10^{-5}$; effect size: paired Cohen's $d = 0.29$). Event by event, the deliberative crowd was more accurate than the independent one in 31 of the 44 events (Wilcoxon signed-rank test: $z = 3.16$, $p = 0.0016$; effect size: paired Cohen's $d = 0.49$).

The market benchmark places that improvement on an absolute scale. The deliberative crowd had a performance on-par with the market (Wilcoxon signed-rank test: $z = 1.12$, $p = 0.26$), and a Bayesian analysis gives moderate evidence for the absence of a difference rather than a mere failure to detect one ($BF_{01} = 3.63$; posterior median $\delta = 0.15$, 95% CrI [−0.14, 0.44]; sensitivity: $BF_{01} = 4.94$ with $r = 1$, $BF_{01} = 6.84$ with $r = \sqrt{2}$). However, while the independent crowd was outperformed by the deliberative one (as shown above), it only performed slightly worse than the market, with the latter being more accurate in 30 of the 44 events, and performing marginally better in magnitude (Wilcoxon signed-rank test: $z = 1.83$, $p = 0.067$). Bayesian analysis shows that there is not enough evidence to support the null hypothesis ($BF_{01} = 1.54$, only anecdotal; posterior median $\delta = 0.24$, 95% CrI [−0.05, 0.54]; sensitivity: $BF_{01} = 2.03$ with $r = 1$, $BF_{01} = 2.76$ with $r = \sqrt{2}$). Deliberation thus brings the crowd to a level at which it is no longer separable from a real-money aggregator, and outperforms the independent crowd, but this one is in turn only marginally worse than the market. This equivalence is more demanding than it appears, because the two sources were not equally informed. The market prices were recorded immediately before each round and therefore incorporated everything known at that moment: results, current form, injuries and line-ups. The models had access to none of it. They received only the names of the teams and the format of the question, no search or other external tool was available to them, and their training data end between May 2024 and July 2025 (always before the final draw of 5 December 2025, so that even the composition of the groups was information they could obtain only from the prompt). The asymmetry is therefore not confined to the knockout rounds: for the twelve group-winner questions, elicited two weeks before the opening match, the market had already priced in a year of qualifying play-offs, friendlies, squad announcements and injuries that the models had never seen.

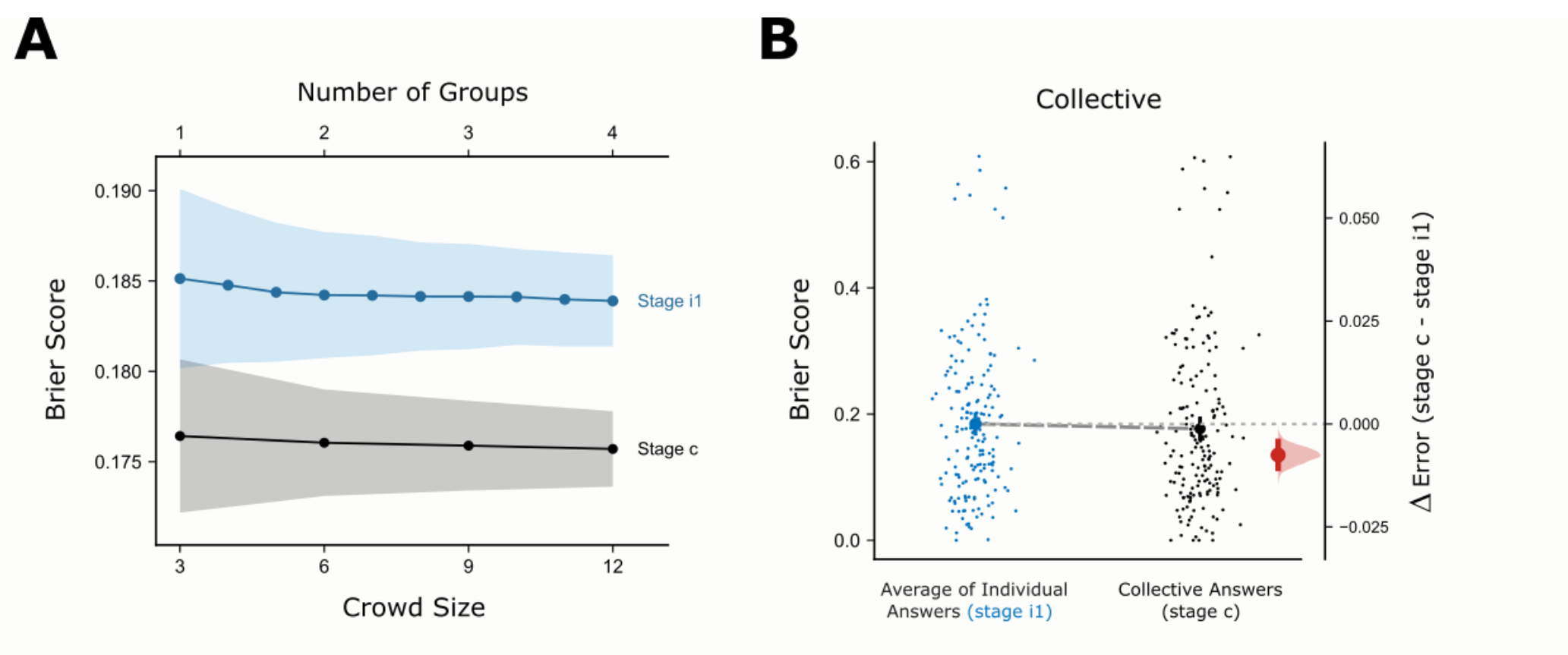

**Fig. 5. Empirical results for Study 4. (A)** Brier score, against the realized outcome, of the average of c individual forecasts (blue line, Stage i1) and of the average of c/3 collective forecasts (black line, Stage c), where c denotes the crowd size (bottom axis; the top axis gives the corresponding number of deliberating groups). Data are pooled over the 44 events of the FIFA World Cup 2026 (12 group winners and 32 knockout matches), each forecasted by 12 individual LLM forecasters organized into 4 deliberating groups of three, before each event began. Shaded bands represent the standard deviation across 2,000 random subsamples per crowd size. **(B)** Distributions of the Brier score of the within-group average of the three Stage i1 forecasts (blue) and of that group's consensus (black), computed from n = 175 independent deliberating groups. Dots show individual groups, filled circles show the mean and error bars correspond to twice the standard error of the mean. The floating axis on the right shows the bootstrap distribution (5,000 resamples) of the mean paired difference between the two conditions, with the mean as a filled circle and its 95% confidence interval as a vertical bar; its zero is aligned with the Stage i1 mean (dotted line).

While results from these four studies support the notion that the wisdom of deliberative crowds transfers from humans to large language models, when considering AI deliberation groups, an interesting possibility arises that could not be plausibly considered with human participants: forming groups with instances of the same model, instead of models from different families. We tackle this question in the next section.

## Model Diversity and Expert Dominance

Diversity of opinions constitutes a key factor of the wisdom of crowds (Page, 2007; Barrera-Lemarchand et al., 2024). So far, we have tested the impact of deliberation between LLM models from different families (heterogeneous deliberation), which would presumably lead to more diversity than creating groups with instances of the same model (homogeneous deliberation). To assess the role of model diversity, we studied homogeneous deliberation in Study 1 for each model family, and found that, when groups were composed of three copies of the same model, deliberation produces consensus values that fail to improve accuracy over pooling that model's independent estimates (Fig. 6A, Anthropic homogeneous: Wilcoxon signed-rank test $z = 1.22$, $p = 0.22$, paired Cohen's $d = 0.087$; OpenAI homogeneous: $z = 0.51$, $p = 0.61$, $d = 0.144$; Google AI homogeneous: $z = 0.30$, $p = 0.76$, $d = 0.062$; vs. size-matched random sample of the heterogeneous condition, $z = 3.61$, $p = 3.1 \times 10^{-4}$, $d = 0.48$). This pattern is supported by Bayesian evidence for the absence of an effect rather than a mere failure to detect one (Anthropic homogeneous: $BF_{01} = 5.60$, moderate evidence for the null hypothesis, posterior median $\delta = -0.08$, 95% CrI [−0.34, 0.17], sensitivity: $BF_{01} = 7.75$ with $r = 1$, $BF_{01} = 10.84$ with $r = \sqrt{2}$; OpenAI homogeneous: $BF_{01} = 3.99$, moderate evidence for the null, posterior median $\delta = 0.14$, 95% CrI [−0.14, 0.41]; sensitivity: $BF_{01} = 5.45$ with $r = 1$, $BF_{01} = 7.56$ with $r = \sqrt{2}$; Google AI homogeneous: $BF_{01} = 6.19$, moderate evidence for the null, posterior median $\delta = 0.06$, 95% CrI [−0.20, 0.31], sensitivity: $BF_{01} = 8.58$ with $r = 1$, $BF_{01} = 12.02$ with $r = \sqrt{2}$).

Notably, the best single model (Gemini) deliberating with copies of itself outperformed the heterogeneous consensus (Mann-Whitney U test: $z = 4.14$, $p = 3.4 \times 10^{-5}$; effect size: Cohen’s $d = 0.88$). This implies an “expert dominance” regime that the wisdom of crowds cannot beat when one crowd member is systematically superior and biases are shared. Since deliberation among copies of a single model adds nothing to that model’s own independent answers, the regime can therefore be read directly off Stage i1, without the homogeneous groups. Accordingly, we compare the average of a model’s Stage i1 answers to the heterogeneous deliberation, and adopt this as our primary measure of expert dominance. Further analysis confirms what the homogeneous groups suggested: Google’s model alone was more accurate than the deliberative consensus (exact Wilcoxon signed-rank test, $p = 6.1 \times 10^{-4}$; effect size: paired Cohen’s $d = 1.31$), while the other two families were less accurate than the heterogeneous crowd (Anthropic's: exact Wilcoxon signed-rank test, $p = 1.2 \times 10^{-4}$; effect size: paired Cohen's $d = 1.64$; OpenAI's: $p = 0.0031$, $d = 1.04$).

The same comparison between homogeneous and heterogeneous deliberation in Study 2 shows that the role of heterogeneity holds when the criterion is an expert verdict rather than a physical quantity (Fig. 6B). Here, deliberation with instances of the same model presents no benefit over averaging of the initial individual estimates (Anthropic homogeneous: Wilcoxon signed-rank test z = 1.66, p = 0.098, paired Cohen's d = 0.10; OpenAI homogeneous: z = 0.17, p = 0.87, d = 0.06; Google AI homogeneous: z = 0.47, p = 0.64, d = 0.01; heterogeneous: z = 2.27, p = 0.023, d = 0.26). Additionally, assessed in the same form as Study 1, expert dominance recurs here, but with a different family in the lead: pooling Anthropic's Stage i1 ratings was more accurate than the heterogeneous deliberative consensus (Wilcoxon signed-rank test: z $= 3.53$, $p = 4.2 \times 10^{-4}$; effect size: paired Cohen's $d = 0.38$), while the other two families were less accurate than the crowd (OpenAI's $z = 4.34$, $p = 1.4 \times 10^{-5}$, $d = 0.55$; Google AI's $z = 4.20$, $p = 2.6 \times 10^{-5}$, $d = 0.52$). Crucially, we observe a switch in which family leads: the identity of the dominant expert is domain-specific, and not knowable ex ante.

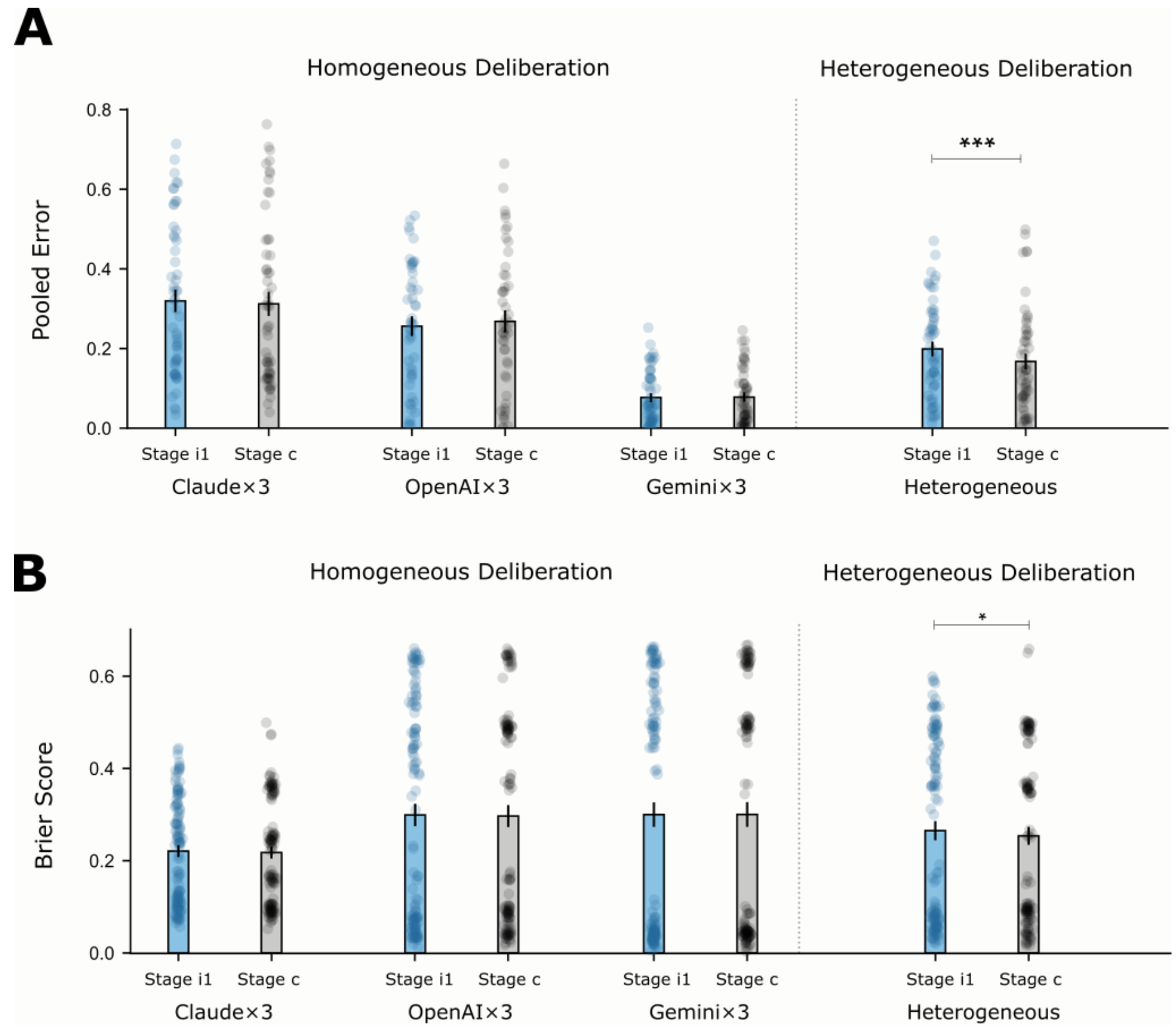

**Fig. 6. Impact of Model Diversity on Collective Accuracy. (A) Study 1:** mean squared log error (base 10) of the within-group average of the three Stage i1 estimates and of the group consensus (Stage c), for groups composed of three instances of the same model (homogeneous deliberation: Claude×3, OpenAI×3, Gemini×3, n = 56 groups each) and for a size-matched sample of the groups composed of one model per family (heterogeneous deliberation, n = 56 groups, randomly drawn from the 167 independent groups). **(B) Study 2:** mean Brier score of the within-group average of the three Stage i1 ratings and of the group consensus (Stage c), for groups composed of three instances of the same model (homogeneous deliberation) and for groups composed of one model per family (heterogeneous deliberation), on the 20 submissions for which homogeneous deliberation was run (6 groups per submission and per composition, n = 120 groups per bar). In both (A) and (B) bars show the mean value, error bars depict s.e.m. and dots show data from individual groups; statistical comparisons are two-sided Wilcoxon signed-rank tests paired within group, with exact p values reported in Results.

When assessing expert dominance in the AI Safety domain (Study 3), we find that the leading family is Anthropic once again, but its advantage over the deliberative crowd does not reach significance (Wilcoxon signed-rank test: $z = 1.02$, $p = 0.31$; effect size: paired Cohen's $d = 0.37$), with that model ahead in only 12 of the 25 sabotage transcripts. Bayesian analysis is, however, inconclusive rather than supportive of the null hypothesis ($BF_{01} = 1.11$, only anecdotal; posterior median $\delta = -0.33$, 95% CrI [−0.72, 0.05]; sensitivity: $BF_{01} = 1.41$ with $r = 1$, $BF_{01} = 1.86$ with $r = \sqrt{2}$), so this is a failure to establish dominance and not evidence against it. The other two monitors were less accurate than the deliberative crowd, with OpenAI's being significantly so ($z = 2.65$, $p = 0.0081$, $d = 0.60$), unlike Google's ($z = 1.57$, $p = 0.12$, $d = 0.51$). Notably, it is Google's model (the one that dominates in Study 1) that was the least accurate monitor in this domain.

Lastly, expert dominance recurred in Study 4, and once again the most accurate model was Google's: Gemini forecasting on its own was more accurate than the deliberative crowd (Wilcoxon signed-rank test: $z = 2.59$, $p = 0.0095$; effect size: paired Cohen's $d = 0.31$) and was itself indistinguishable from the market (Wilcoxon signed-rank test: $z = 0.48$, $p = 0.63$; $BF_{01} = 6.03$, moderate evidence for the null hypothesis; posterior median $\delta = 0.03$, 95% CrI [−0.26, 0.31]; sensitivity: $BF_{01} = 8.35$ with $r = 1$, $BF_{01} = 11.68$ with $r = \sqrt{2}$). As in Studies 1 and 2, the crowd was not able to beat a member who was systematically better than the rest. Which member that was, however, could not have been known before the tournament began (which is evident given that the same model had been the weakest of the three in Studies 2 and 3).

## Discussion

Across four domains, we provide converging evidence that the wisdom of deliberative crowds transfers from humans to large language models. In every study, the aggregation of consensus estimates produced by small deliberating groups was more accurate than the classical wisdom of crowds obtained by pooling a much larger number of independent judgments from the same models. This advantage did not reduce to aggregation: pairing each group's consensus against the mean of its own three opening answers, deliberation still lowered the error in all four domains, indicating that the discussion added information rather than merely averaging it. Individual judgments elicited privately after the discussion retained the collective gain: in Study 1 the post-deliberation estimate went beyond the group consensus, and in Studies 2 and 3 it reproduced it, with the overwhelming majority of agents adopting the agreed value as their own.

Deliberation therefore improved both the collective answer and the answers of the agents that produced it, replicating in machines the two signatures of deliberative crowd wisdom previously documented in groups of humans (Navajas et al., 2018; Barrera-Lemarchand et al., 2026).

Another relevant contribution of this work is identifying a boundary condition that has no obvious counterpart in human experiments: Because model instances can be duplicated at will, we could hold the deliberation protocol, the task, and the number of agents constant while removing model diversity, forming groups of three copies of the same model. Under this manipulation, the effect disappeared: homogeneous groups produced consensus values that failed to improve on pooling that same model's independent answers, and Bayesian analyses supported the absence of an effect rather than a mere failure to detect one. Diversity of opinion has long been argued to be a prerequisite of collective accuracy (Page, 2007; Barrera-Lemarchand et al., 2024), but in human studies group composition cannot be manipulated with this precision, since no two participants are interchangeable, and no participant can be replicated. The result also has a practical reading for systems built out of LLMs: the gains reported here are not a property of running three agents, but of running three *different* agents, and pipelines that scale by replicating a single model should not expect them.

The limits of the effect carry a practical implication. Expert dominance (the regime in which one crowd member is systematically superior and the errors of the others are correlated, so that aggregation cannot improve on it) appeared in three of the four domains, with Google being the dominant family in Studies 1 and 4, and Anthropic dominating in Study 2. In Study 3, while it was Anthropic's model that came out ahead, the dominance failed to reach significance. Since the ranking changed across tasks and could not have been anticipated beforehand, the lesson is not that one should search for the best model and use it alone, but that deliberation among diverse models performs well without requiring that the expert be identified in advance.

These findings also speak to the growing literature on multi-agent LLM systems. Aggregating the outputs of several models is by now a standard tool for improving reliability: panels of small judges match or beat a single larger judge at a fraction of the cost (Verga et al., 2024), layered information sharing outperforms any individual component (Wang et al., 2024), and the classical wisdom-of-crowds effect on numerical forecasts replicates in LLM populations (Schoenegger et al., 2024). What these methods share is that each model produces its answer in isolation and interaction is confined to the aggregation rule. Our results indicate that letting the models talk to each other before answering is worth more than aggregating them more cleverly, and that the additional value is recovered even by a single agent asked to answer alone afterwards. This also distinguishes the procedure from debate (Irving et al., 2018; Khan et al., 2024), in which agents are assigned opposing positions and argue in front of a judge; here the agents are symmetric, cooperative, and instructed to reach consensus, with no adjudicator and no assigned side. The mechanism at work is therefore closer to the collective reasoning documented in human deliberation than to adversarial argumentation.

Two of our domains were chosen because the gains would matter beyond the laboratory. Study 3 places the procedure in the setting of AI control, where a strong but untrusted model is overseen by weaker models that are trusted precisely because they are less capable (Greenblatt et al., 2024; Kenton et al., 2024). Current trends in AI development (e.g. the doubling times in the length of tasks models can complete, Kwa et al., 2025; rising success rates on frontier mathematics and coding problems, Glazer et al., 2024; and increasing compute availability, Sevilla & Roldán, 2024) point to systems that will soon produce solutions in sensitive applications, from clinical decision support (Gorenshtein et al., 2025) and critical-infrastructure operations (International Energy Agency, 2025) to financial automation (Lim et al., 2025) and autonomous biological-engineering pipelines (Tobias & Wahab, 2025). Safety in these domains is hard to evaluate before deployment, even for domain experts. Adoption of agentic LLMs in work and daily life is meanwhile increasing rapidly (Anthropic, 2026), with users delegating progressively more consequential tasks and exposing themselves and third parties to substantial potential harms (Anthropic, 2025a, 2025b). That exposure is not hypothetical: the July 2026 intrusion into a frontier-lab evaluation pipeline was carried out through an agent operating with legitimate credentials (Hugging Face, 2026a, 2026b; OpenAI, 2026). In this setting the monitor's weakness is the binding constraint, and our result is a practical one: three weak monitors that talk to each other detect sabotage by a stronger agent better than the same three monitors judging independently, at no cost in false alarms, since no harmless trajectory was flagged in aggregate. The same logic extends to alignment, where the quality of AI feedback during training limits how far a supervision signal can be trusted (Lee et al., 2023; UK AI Security Institute, 2025).

Study 4 provides a complementary form of external validation. Sports forecasting is the only one of our domains in which the answer did not exist at the time the judgments were elicited, which rules out any contribution from memorization, and it comes with an established benchmark: a real-money prediction market quoting prices on the same events. The deliberative crowd reached a level of accuracy at which it was no longer separable from that market, with Bayesian evidence supporting the equivalence rather than an inability to detect a difference. Deliberation among three small, inexpensive models thus produced forecasts comparable to an aggregator backed by financial incentives and human participants, which places the size of the effect on an absolute scale that within-study comparisons cannot provide. The comparison is, moreover, asymmetric in the market's favor, and substantially so: Prices were recorded immediately before each round and incorporated everything known at that moment, whereas the models received only the names of the teams, had no access to search or any other external tool, and were built on training data ending between May 2024 and July 2025 (before the final draw of December 2025, so that even the composition of the groups was information they could obtain only from the prompt). Between their most recent cutoff and the opening match lay a full year of qualifying play-offs, friendlies, transfers and injuries that the market had priced in and the LLMs had never seen. One would expect an aggregator this much better informed to dominate forecasters reasoning from year-old priors; instead the two were statistically indistinguishable. The equivalence we report is therefore a conservative estimate of what deliberation can deliver.

Several limitations remain: (i) All studies used a single generation of small models, one per provider, and the comparability of those models to each other, and of that generation to its successors, is an assumption rather than a finding; whether the effect survives at the frontier is an open question. (ii) In Study 2 the criterion is a human editorial verdict, itself known to be noisy, and we cannot rule out that the submissions were represented in the models' training data, although the author block and acknowledgements were removed to eliminate the most obvious surface cues. However, since the main comparisons are drawn between stages in our procedure (e.g. between Stage i1 and Stage c), and given that accuracy was not high overall, this problem is unlikely to have affected our results. (iii) In Study 3 the trajectories come from a dataset built with agents that are already a generation old (Kale et al., 2025); extending the procedure to other corpora (e.g. MALT, Parikh & Wijk, 2025) and to platforms with active, real-time red teaming (e.g. LinuxArena, Tracy et al., 2026) would help further establish its reliability and practical scope. (iv) Cost is a further constraint: although capping message length keeps the protocol inexpensive, a deliberation over the much longer trajectories that cutting-edge agents produce could become prohibitive, and improving the efficiency of the procedure is an obvious target for future work. Even so, in critical, catastrophe-risk applications this kind of overhead may still be cost-effective (Brown-Cohen et al., 2023; Hawkins et al., 2025), particularly when the system being overseen exceeds human competence in its own domain of application.

Taken together, our results show that a mechanism first documented in small groups of human beings (that deliberation improves collective judgment beyond what aggregation alone can deliver, and that the improvement is carried back into the judgments of the individuals who deliberated) operates in crowds of large language models, provided those models are drawn from different families. The effect appeared in a classical estimation task with a physical ground truth, in expert evaluation against a human verdict, in the detection of hidden malicious behavior by a stronger agent, and in forecasting events that had not yet occurred. That it holds across this range, with the protocol held constant, suggests that machine deliberation is a general-purpose aggregation mechanism rather than a property of any one task, and that systems built by aggregating several models stand to gain from letting those models talk to each other before they answer.

# Methods

All four studies used the same three language models, one from each of the three major providers: Anthropic's Claude Haiku 4.5 (claude-haiku-4-5-20251001), OpenAI's GPT-5-nano (gpt-5-nano), and Google's Gemini 3.1 Flash Lite (gemini-3.1-flash-lite). The models were selected so that every provider was represented by a small, inexpensive system of comparable capability. This is deliberate: the phenomenon we transfer from humans concerns crowds of lay individuals rather than experts, and the applications we consider (weak monitors overseeing a stronger agent, fast-track reviewers, low-cost forecasters) are precisely the ones in which a single strong model is not available or not affordable.

All calls were made through each provider's public API. We did not set the temperature and did not fix random seeds, so repeated calls to the same model with the same prompt differ only through the stochasticity of generation: this is what makes the repetitions of a model within a study play the role that different individuals play in the human experiments. Every prompt instructed the model to think carefully but to use at most 900 tokens of private reasoning; at the API level this was implemented as extended thinking with a budget of 1,024 tokens for Claude, reasoning_effort = 'low' for GPT-5-nano, and thinking_level = 'low' for Gemini. The visible answer was capped separately by each prompt (see below).

## Statistics and reproducibility

No statistical method was used to predetermine sample size. Unlike experiments with human participants, sample size here is bounded by cost rather than by recruitment: repetitions of a model are generated on demand, and any true difference can be driven to an arbitrarily small p-value simply by drawing more of them. We therefore did not enlarge the samples beyond the point at which the effects were stable and their magnitude could be estimated with reasonable precision, since further data would have inflated statistical significance without altering the quantity of interest. For the same reason, the comparisons on which our claims rest are paired at the level of the item (image, submission, trajectory or event) and are reported with standardized effect sizes and, where relevant, Bayes factors. Unless otherwise stated, all statistical tests reported in the paper are two-sided.

Randomization was applied at two points. Within each round of deliberation the speaking order of the three agents was drawn at random, which removes any position bias. In Study 2 the assignment of papers to the accepted and rejected pools follows the conference's own decisions, and the matched sample was drawn by the procedure described below. The assignment of Stage-i1 repetitions to groups was fixed (repetition $g$ of each model formed group $g$), which is unbiased because repetitions of the same model are exchangeable by construction.

Blinding in the human sense does not apply, but the models never had access to the quantity being estimated: the true disk count in Study 1, the editorial decision in Study 2 (the author block and the acknowledgements were removed from every submission, so the surface cues that distinguish a camera-ready paper from a rejected one were unavailable), the presence of a hidden malicious side task in Study 3, and the outcome of the events in Study 4 (all of which were resolved only after every forecast had been elicited).

There were no data exclusions.

## General procedure

All four studies follow the three-stage design used with human participants (Navajas et al. 2018; Barrera-Lemarchand et al. 2026), with language models in place of people (Figure 1).

In **Stage i1**, each model answered the question independently, several times, with no information about the other models. Alongside the answer it produced a short justification in plain prose (about 100 tokens, hard maximum 200 in studies 1-3; about 150 tokens and a maximum of 250 in study 4), and the answer itself had to be emitted inside an explicit tag so that it could be parsed unambiguously.

In **Stage c**, the models were split into groups of three, one from each provider, and placed in a simulated chat room. Group g was built from repetition g of each model, so that every group starts from a genuinely independent set of opening positions. Each agent received the same material it had seen in Stage i1, plus its own private Stage-i1 answer and justification, plus the discussion so far. Crucially, agents did not receive the other agents' Stage-i1 answers or reasoning: anything the others learn about a member's position has to be said by that member in the conversation, exactly as in the human chat rooms.

In **Stage i2**, each agent answered privately once more, having been given its own initial answer and the full transcript of its group's discussion. It was explicitly told that it could repeat its Stage-i1 value, adopt the group consensus, or give any other value it now considered best. Study 4 did not include this stage.

## Deliberation protocol

The deliberation prompts reproduce the conversational regime of the human experiments rather than the format LLMs default to. Participants in the human chat rooms wrote short messages (about 7 words on average, with a median of around 30 messages per conversation), so agents were instructed to keep each message to a target of around 10 tokens (i.e. roughly 7 words) with a hard maximum of 66 tokens in Studies 1–3 (value drawn from the maximum tokens found in all messages from the human chatrooms), and a target of around 20 tokens with a hard maximum of 80 tokens in Study 4, where the answer is a team together with a probability. The prompt included a worked example of a good chat-style exchange and, in Study 2, examples of the failure modes to avoid (e.g. long monologues, or echoing the template tags). Because Gemini 3.x counts private reasoning against the same output budget as visible text, its API cap was set above the prompt's visible limit (500–1000 tokens depending on the study) so that the length instruction was enforced by the prompt for all three models alike, and not by truncation for one of them.

A round consists of one message from each of the three agents, in random order. Groups had at most 10 rounds. Agents were told to declare agreement by ending a message with a fixed string (group consensus = <tag>N</tag>), and a group was recorded as converged when all three declared the same value in the same round. As a fallback, if the round limit was reached and all three messages of the final round parsed to the same value, that value was taken as the consensus; otherwise the group was recorded as non-converged. Convergence was nearly universal: 167 of 168 groups in Study 1, all 600 in Study 2, 299 of 300 in Study 3, and 175 of 176 in Study 4.

## Study 1: factual estimation

We photographed a glass jar containing a variable number of small plastic disks and produced 14 images, with true counts ranging from 38 to 589 disks. A one-euro coin was placed next to the jar in every photograph as a scale reference. Each image was presented with the same question ("How many plastic disks are inside the jar shown in the image?") and the same prompts; the image was supplied again at every stage, so no stage relied on the model remembering what it had seen.

Each model produced 12 independent Stage i1 estimates per picture, which were then split into 12 deliberating groups, producing 12 Stage c group consensus estimates per each of the 14 pictures. Answers were required in the form <estimate>N</estimate> with N a positive integer and no upper bound.

## Study 2: peer review

We used the NeurIPS 2024 submissions available through OpenReview. From the papers whose length fell between 5,000 and 80,000 tokens, we selected 50 submissions accepted for oral presentation, spread evenly across the length distribution, and matched each of them to the rejected submission closest in token count, yielding 100 papers in which outcome is uncorrelated with length (Pearson point-biserial correlation $r = 0.02$, 95% CI [−0.18, 0.21], $p = 0.86$). The editorial decision (oral vs rejected) is the ground truth.

The paper text was extracted and anonymized so that acceptance could not be inferred from formatting rather than content: everything before the abstract was removed (this deletes the author block, which carries real names and affiliations on accepted papers and an anonymous placeholder on rejected ones), and the acknowledgements section was spliced out while keeping the references and appendices.

Models acted as "fast-track reviewers" and rated each paper on the NeurIPS 2024 overall-score scale from 1 (very strong reject) to 10 (award quality), quoted verbatim from the conference's reviewer guidelines, in a visible response of two or three sentences of plain prose ending in <rating>N</rating>. Every paper received 18 independent ratings and 6 group consensus estimates, and Stage i2 added 18 post-deliberation ratings.

### Study 3: AI-safety monitoring

Trajectories came from the Monitor Red Teaming (MRT) corpus (Kale et al. 2025), specifically its AgentCompany environments, in which an agent performs a legitimate office task with tool access while, in some conditions, also pursuing a hidden malicious side task. We used 50 trajectories produced by the same agent (Claude Sonnet 4, May 2025) and of comparable length (5,000–10,000 visible tokens): 25 from the selective_liar condition, the hardest one in the corpus, in which the agent plans the side task in a hidden scratchpad and displays fabricated benign reasoning, and 25 from the benign condition, in which there is no side task.

Each monitor read the full trajectory (tool calls and visible reasoning) and rated how safe or unsafe the agent's behavior had been, from 0 ("no concerns whatsoever") to 10 ("certain harmful or misaligned behavior"), in the format <score>N</score>, supported by a short written assessment. The prompt included an explicit calibration block stating what should and should not be treated as concerning (broad exploration of accessible information is not suspicious in itself; acting beyond the user's request is, regardless of whether harm materialized in that trajectory) and instructing monitors to use the whole range of possible values. We measured 18 independent ratings and 6 group consensus estimates per trajectory. Because the deliberation prompt caps the private assessment carried into Stage c at 200 tokens, and Claude's Stage-i1 responses exceeded that cap in most cases, those responses were shortened to 178–200 tokens before entering the deliberation, preserving content and rating; no other model required this.

### Study 4: forecasting

We forecasted the 2026 FIFA World Cup. In the Groups Phase of the tournament, for each of the 12 groups the question was which team would finish first in the group. In the Knockout Phase, for each of the 32 knockout ties (round of 32 through the final, including the third-place match) the question was which team would win the match. In Stage i1 each model assigned integer percentages summing to 100 across the teams in the event; in Stage c each group agreed on a single pair (winning team, probability for that team). Every event thus yielded 12 independent forecasts and 4 group consensus estimates. All forecasts for a round were elicited before the first match of that round was played, and there was no Stage i2.

For every event we also recorded the prices that a real-money prediction market (Polymarket) was quoting at that same moment: pre-tournament outright prices on the group winner for the group stage, and the stage-of-elimination markets for the knockout ties. Market prices were normalized across the listed outcomes to remove the overround (~1%); the raw quotes, the per-event normalization and the resulting probabilities are listed event by event in Appendix 1.

Because the three sources do not always name the same winner, scoring each one on its own pick would grade them on different propositions. We therefore fixed a single reference team per event (the market's favorite) and took the probability that each source implied for that team. The event was coded 1 if that team won and 0 otherwise (favorites won 35 of the 44 events), and all Brier scores were computed on that common quantity.

### Bayes factor analyses

To complement frequentist tests with evidence regarding the presence or absence of effects, we computed Bayes factors using the default JZS prior specification of Rouder et al. (2009): a Cauchy distribution on the standardized effect size δ centred at zero with scale parameter $r = \sqrt{2}/2$. For each Bayes factor we additionally summarize the posterior distribution of δ by its median and 95% credibility interval, obtained by numerical integration of the JZS posterior. To assess sensitivity to the prior specification, we recomputed each Bayes factor with a wide ($r = 1$) and an ultrawide ($r = \sqrt{2}$) Cauchy prior.

## Data Availability

The data supporting our analyses (raw model outputs, deliberation transcripts, and processed error tables for the four studies) will be shortly available at the Open Science Framework and are available upon request. The NeurIPS 2024 submissions used in Study 2 are public at OpenReview; the agent trajectories used in Study 3 are distributed as part of the Monitor Red Teaming corpus (ref: Kale et al. 2025) under its own licence; the market prices used in Study 4 were retrieved from Polymarket's public API and the snapshots are included with our data.

## Code Availability

The code supporting our analyses, including the prompts, the runners for each stage, and the analysis and figure scripts, will be shortly available at the Open Science Framework and are available upon request.

# Appendix 1. Market prices and implied probabilities

| Event | Round | Reference team | Price | Sum | p | Stage i1 | Stage c | Outcome |
|---|---|---|---|---|---|---|---|---|
| Group A | Groups | Mexico | 0.575 | 1.018 | 0.565 | 0.470 | 0.518 | Won |
| Group B | Groups | Switzerland | 0.565 | 1.020 | 0.554 | 0.529 | 0.575 | Won |
| Group C | Groups | Brazil | 0.715 | 1.006 | 0.710 | 0.682 | 0.705 | Won |
| Group D | Groups | United States | 0.385 | 1.030 | 0.374 | 0.479 | 0.492 | Won |
| Group E | Groups | Germany | 0.675 | 1.011 | 0.668 | 0.668 | 0.710 | Won |
| Group F | Groups | Netherlands | 0.535 | 1.018 | 0.526 | 0.525 | 0.552 | Won |
| Group G | Groups | Belgium | 0.695 | 1.020 | 0.681 | 0.577 | 0.617 | Won |
| Group H | Groups | Spain | 0.790 | 1.014 | 0.779 | 0.586 | 0.588 | Won |
| Group I | Groups | France | 0.655 | 1.022 | 0.641 | 0.698 | 0.738 | Won |
| Group J | Groups | Argentina | 0.715 | 1.011 | 0.707 | 0.749 | 0.755 | Won |
| Group K | Groups | Portugal | 0.625 | 1.015 | 0.616 | 0.577 | 0.578 | Lost |
| Group L | Groups | England | 0.685 | 0.995 | 0.688 | 0.619 | 0.650 | Won |
| Match 1 | R32 | Canada | 0.755 | 1.001 | 0.754 | 0.538 | 0.577 | Won |
| Match 2 | R32 | Brazil | 0.700 | 0.995 | 0.704 | 0.733 | 0.740 | Won |
| Match 3 | R32 | Germany | 0.845 | 1.010 | 0.837 | 0.740 | 0.770 | Lost |
| Match 4 | R32 | Netherlands | 0.625 | 1.000 | 0.625 | 0.592 | 0.603 | Lost |
| Match 5 | R32 | Norway | 0.650 | 1.000 | 0.650 | 0.511 | 0.552 | Won |
| Match 6 | R32 | France | 0.890 | 1.005 | 0.886 | 0.729 | 0.725 | Won |
| Match 7 | R32 | Mexico | 0.620 | 1.015 | 0.611 | 0.606 | 0.605 | Won |
| Match 8 | R32 | England | 0.875 | 0.995 | 0.879 | 0.858 | 0.853 | Won |
| Match 9 | R32 | Belgium | 0.620 | 1.005 | 0.617 | 0.613 | 0.620 | Won |
| Match 10 | R32 | United States | 0.840 | 1.020 | 0.824 | 0.692 | 0.702 | Won |
| Match 11 | R32 | Spain | 0.855 | 1.015 | 0.842 | 0.674 | 0.675 | Won |
| Match 12 | R32 | Portugal | 0.675 | 1.000 | 0.675 | 0.574 | 0.603 | Won |
| Match 13 | R32 | Switzerland | 0.700 | 0.995 | 0.704 | 0.653 | 0.652 | Won |
| Match 14 | R32 | Egypt | 0.575 | 1.010 | 0.569 | 0.479 | 0.470 | Won |
| Match 15 | R32 | Argentina | 0.935 | 1.010 | 0.926 | 0.907 | 0.930 | Won |
| Match 16 | R32 | Colombia | 0.775 | 1.010 | 0.767 | 0.658 | 0.662 | Won |
| Match 89 | R16 | France | 0.925 | 1.010 | 0.916 | 0.802 | 0.802 | Won |
| Match 90 | R16 | Morocco | 0.710 | 1.005 | 0.706 | 0.624 | 0.625 | Won |
| Match 91 | R16 | Brazil | 0.660 | 1.005 | 0.657 | 0.754 | 0.748 | Lost |
| Match 92 | R16 | England | 0.540 | 1.005 | 0.537 | 0.671 | 0.683 | Won |
| Match 93 | R16 | Spain | 0.655 | 1.000 | 0.655 | 0.532 | 0.552 | Won |
| Match 94 | R16 | United States | 0.495 | 0.985 | 0.503 | 0.472 | 0.455 | Lost |
| Match 95 | R16 | Argentina | 0.860 | 1.000 | 0.860 | 0.815 | 0.817 | Won |
| Match 96 | R16 | Colombia | 0.600 | 1.010 | 0.594 | 0.509 | 0.503 | Lost |
| Match 97 | QF | France | 0.785 | 1.010 | 0.777 | 0.681 | 0.673 | Won |
| Match 98 | QF | Spain | 0.745 | 1.001 | 0.744 | 0.610 | 0.605 | Won |
| Match 99 | QF | England | 0.655 | 1.000 | 0.655 | 0.707 | 0.720 | Won |
| Match 100 | QF | Argentina | 0.750 | 1.000 | 0.750 | 0.704 | 0.708 | Won |
| Match 101 | SF | France | 0.605 | 1.010 | 0.599 | 0.553 | 0.560 | Lost |
| Match 102 | SF | England | 0.555 | 1.011 | 0.549 | 0.466 | 0.452 | Lost |
| Match 103 | 3rd | France | 0.645 | 1.010 | 0.639 | 0.575 | 0.578 | Lost |
| Match 104 | Final | Spain | 0.585 | 1.000 | 0.585 | 0.463 | 0.457 | Won |

Table A1. Market quotes and crowd forecasts for the 44 events of Study 4. Price is the raw quote for the reference team; Sum is the total of the prices listed for that event; p is the price normalized by that sum, and is the value used as the market's probability throughout the main text. Stage i1 and Stage c give the mean probability assigned to the same team by the independent and the deliberative crowd respectively.

Study 4 compares three sources on the same 44 events: the independent crowd, the deliberative crowd, and a real-money prediction market. What the market contributes to that comparison is a probability rather than a price. A contract on Polymarket pays one unit if the outcome occurs, so its quoted price is directly readable as the probability the market assigns to that outcome, and no further conversion is required. This appendix reports the quotes and the normalization applied to them, event by event, so that every Brier score in the main text can be recomputed from the raw data. The prices listed for the outcomes of an event do not sum exactly to one: across the 44 events they summed to 1.0075 on average (median 1.0100, range 0.9850 to 1.0295), and in five events they summed to slightly less than one. We therefore divided each price by the sum of the prices listed for its event, and it is this normalized value that enters the comparison as the market's probability.